**Assessing mortality prediction through different representation models based on concepts extracted from clinical notes**

**Hoda Memarzadeh**

Department of Electrical and Computer Engineering, Isfahan University of Technology, Isfahan 84156-83111, Iran. Email: h.memarzadeh@ec.iut.ac.ir

**Nasser Ghadiri [1]**

Department of Electrical and Computer Engineering, Isfahan University of Technology, Isfahan 84156-83111, Iran. Email: nghadiri@iut.ac.ir

**Maryam Lotfi Shahreza**

Department of Computer Engineering, Shahreza Campus, University of Isfahan, Iran. Email: m.lotfi@shr.ui.ac.ir

*Abstract*

Recent years have seen particular interest in using electronic medical records (EMRs) for secondary purposes to enhance the quality and safety of healthcare delivery. EMRs tend to contain large amounts of valuable clinical notes. Learning of embedding is a method for converting notes into a format that makes them comparable. Transformer-based representation models have recently made a great leap forward. These models are pre-trained on large online datasets to understand natural language texts effectively. The quality of a learning embedding is influenced by how clinical notes are used as input to representation models. A clinical note has several sections with different levels of information value. It is also common for healthcare providers to use different expressions for the same concept. Existing methods use clinical notes directly or with an initial preprocessing as input to representation models. However, to learn a good embedding, we identified the most essential clinical notes section. We then mapped the extracted concepts from selected sections to the standard names in the Unified Medical Language System (UMLS). We used the standard phrases corresponding to the unique concepts as input for clinical models. We performed experiments to measure the usefulness of the learned embedding vectors in the task of hospital mortality prediction on a subset of the publicly available Medical Information Mart for Intensive Care (MIMIC-III) dataset. According to the experiments, clinical transformer-based representation models produced better results with getting input generated by standard names of extracted unique concepts compared to other input formats. The best-performing models were BioBERT, PubMedBERT, and UmlsBERT, respectively.

---

[1] Corresponding author. Altername email: nghadiri@gmail.com

## 1    Introduction

A patient's Electronic Medical Record (EMR) includes specific information about their health history, e.g., diagnoses, medicines, tests, allergies, immunizations, and treatments. EMRs are digital files containing both structured and unstructured data.

Structured data are 'recorded in a patient's file using standard coding systems such as laboratory tests and medications records. Examples of such data are diagnostic codes or prescriptions and laboratory tests recorded using standard terminology systems, such as ICD10[2], RxNorm[3], or LOINC[4].

In contrast, unstructured data are the free texts written by service providers in various sheets. The unstructured data stored textually in an EMR contains valuable information such as the patient's history or explanations about their illness when entering the hospital. However, processing these texts could be difficult because of challenges such as more syntactic differences, word choices, abbreviations, acronyms, compounds, and negation expressions, as well as speculative cues, factuality expressions, and spelling errors compared to general notes (1). The complexity of processing unstructured data has turned it into one of the most researched topics.

In this article, we focus on how to model unstructured data to make it usable as the input of patient information representation models. Two steps, i.e., named entity recognition (NER) and named entity disambiguation or entity linking, are necessary to overcome the challenges of unstructured data, which include clinical texts recorded by healthcare providers.

The NER step identifies and aligns biomedical entities to knowledge bases, enhancing decision-making efficiency (2) and reducing the diversity of clinical terms. A key goal of entity disambiguation and entity linking is to give each entity in the text a unique identifier. Through entity linking, entities extracted from the text can be mapped to corresponding unique ids from a knowledge base such as the Unified Medical Language System (UMLS). The problem that arises in the application of NER and entity linking methods is the extraction of concepts. The extracted concepts can be repetitive and may not have the same differential value.

---

[2] International statistical classification of diseases and related health problems
[3] Normalized naming system for generic and branded drugs
[4] Logical Observation Identifiers Names and Codes

In this study, during the section detection step, the initial discharge sheet was split into inner sections, and after filtering more valuable sections by performing the NER process, their included notes were converted to a list of concepts mapped with the knowledge base. The produced set was very likely to contain repetitive and insignificant concepts. This research's proposed input was designed based on a unique subset of concepts extracted from each file. The extracted unique concepts were used as input to ten different textual representation methods. After the mentioned processing steps, the impacts of input type and representation model type on the final result were examined from different aspects.

In this article, we investigated how the use of external knowledge to process clinical notes could affect the quality of representation models. In summary, the contributions of this work are as follows:

1. The effect of exploiting external knowledge in the form of medical concepts' standard names as input was investigated.
2. The performance of the representation models, which received the input generated by the proposed method, was evaluated. The representation models were pre-trained on different medical vocabularies and knowledge sources.
3. The performance of different clinical representation models using different processing methods on three types of clinical notes as input was also studied.

The rest of this paper is organized as follows. Section 2 introduces the proposed framework for generating representations from different versions of text inputs and explains how the prediction model is created. The results of several organized models are examined in Section 3. Finally, Section 4 discusses the results and presents the conclusions.

## 1.1    Statement of significance

| Problem or Issue | A number of challenges are associated with converting clinical texts into comparable vectors that can be used in machine learning tasks. Clinical texts in EMR differ in their expression of specialized expressions, their writing styles, and the importance of different sections. |
|---|---|

| What is Already Known | Despite the many advances made in document representation by transformer-based representation models, the input type of these models still affects the quality of the generated embedding vector. |
|---|---|
| What this Paper Adds | In this article, we identify the different sections of the discharge sheet, one of the most richly detailed texts in the EMR. After selecting a few more important sections for predicting a patient's outcome, their textual content is processed with the UMLS knowledge base by Spacy-based libraries. As a result of removing the redundant concepts, a summary text consisting of standard expressions was produced, which was used as input for clinical transformer models. When this input is used for the in-hospital mortality prediction task, the constructed embedding vectors show a quality improvement. |

## 2    Related works

A review of research on the use of machine learning methods on electronic file data shows that it is crucial to pay attention to both the type of input and type of representation model and design architecture to solve the downstream task.

### 2.1    Input data type

Regarding input data type, a review of articles on EMR representation shows that initially, only structured data recorded in electronic files were used as input to algorithms. For example, using structured information representations, Choi et al. predicted heart failure using diagnoses recorded during patient visits (3,4). In such studies, the information was represented using word2vec, which is a neural network-based method (5,6). Mittio et al. (7) showed that using unstructured data, along with structured data, improved the results reported in (3,4). They used the auto-encoder method for representation, and given the fact that using this method was reported to have no significant effect on the results in (3), the obtained improvement can be attributed to the use of textual data. Employing the transformer model, Dligach et al. showed that using unstructured data alone could provide comparable results with those of other methods that use both unstructured and structured data  simultaneously (8).

### 2.2    NLP in clinical notes

A key step in natural language processing (NLP) is entity recognition. This process can be rule-based or a combination of methods in which machine learning models are used to reinforce rules (9). In a review study (10) published in 2016, no difference was reported between rule-based methods and those based on machine learning. Still, in another review study conducted in 2019, the share of the machine learning method witnessed a significant growth (11). The rule-based method requires humans to handcraft specific rules to identify different types of named entities; In contrast, machine learning approaches automatically learn patterns for identifying named entities based on a large corpus of labeled training data. The main advantage of machine learning approaches over rule-based ones is their higher flexibility. However, they require large quantities of word-level annotations, which are costly and impossible to obtain within biomedical settings (12). MetaMap (12) and cTAKES (13) are examples of the first group, and BioBERT (14) and CLAMP(15) are in the second group (16).

ScispaCy is an instance of the tools that employ the rule-based technique. It is a Python-based NLP pipeline developed by the Allen Institute for Artificial Intelligence (AI2) designed to analyze scientific and biomedical text using natural language processing (17). It includes a custom tokenizer that adds tokenization rules onto SpaCy's rule-based tokenizer (18), a POS tagger and syntactic parser trained on biomedical data, and an entity span detection algorithm.

The extracted entities were mapped to the concepts defined in the ontology to reduce ambiguity. This was done in the entity linking stage. With entity linking, extracted entities from the text were mapped to corresponding unique ids from a target knowledge base. The target knowledge base in MetaMap and cTAKES is UMLS. In Scispacy, the target knowledge base could be UMLS, MeSH, RxNorm, GO, or HPO (17). This step has been made for various purposes in previous studies. For example, in (19), the definitions in Metatarsus were used to improve the quality of embedding vectors generated for concepts. Moreover, in (20), the concepts mapped to UMLS were used to design a model for identifying heart problems. In (21), the extracted concepts mapped with SNOMED-CT ontology were used to make connections between patients and treatment methods.

Despite the added value achieved through NER implementation and linking, there is always the issue that the set of concepts extracted (bag of concepts) includes concepts that do not have differentiating characteristics. In other words, more valuable concepts need to be separated from other extracted concepts. However, one of the research topics is how to use the extracted concepts and identify the key concepts.(22,23).

Identifying key concepts in clinical notes is more sensitive because these notes are composed of sections that address various issues related to the patient treatment process. It is important to identify these sections before identifying the concepts. Depending on which section is analyzed, a clinical condition may have been experienced by the patient in the past, may be ongoing at the time of the note, or the patient may be at risk of experiencing it in the future. Depending on the section type, there might be a variety of section headers spelled out or abbreviated. Accordingly, section detection is a core part of clinical NLP. MedSpaCy implements clinical section detection through rule-based matching of section titles, adapted from the default SecTag rules (24).

## 2.3 Representation models

Regardless of the importance of using textual data and how to process them, the clinical data in machine learning problems should be converted into a comparable format, which is done by learning embedding vectors. Over the years, researchers have developed numerous text representation techniques, which started with one-hot and bag-of-words models and expanded to RNN-based and transformer-based models. Given the general success of transformer-based pre-trained language models, many such models have been developed by pre-training on biomedical corpora. The following is a quick review of some of these models used in this paper. The BERT-BASE (11) model has been trained on the Wikipedia and book corpus databases but has not been fine-tuned on clinical data and can therefore be considered a general model. The BioBERT architecture uses the weights of BERT-BASE but has been trained on PubMed abstracts and PubMed Central (PMC) archive, in addition to the databases mentioned above (12). Using the original vocabulary of BERT-BASE in BioBERT allows for reusability of previously created BERT-BASE models and interchangeability between BERT-BASE and BioBERT. The model named BlueBERT has used clinical notes in the MIMIC-III dataset in addition to PubMed abstracts (13). However, another model called Bio+Clinical BERT has been only trained on clinical notes in the MIMIC-III dataset (14). Unlike the above models, which all use the vocabulary of BERT-BASE, the model called SciBERT has its own vocabulary, which is about the same size as that of BERT-BASE. This vocabulary contains 30,000 sub-words and has been compiled from 1.14 million articles (15). Another model that has its own vocabulary is PubMedBERT. This vocabulary, which contains more words specific to the biomedical field than SciBERT, has been compiled from 30 million PubMed abstracts (16). The model named UmlsBERT has been developed based on the knowledge augmentation strategy and the learning of UMLS semantic groups such as anatomy and disorder (17). Although BlueBERT and Bio+Clinical BERT were also trained on clinical data, the ability to learn semantic groups is a unique feature of UmlsBERT.

## 3    Materials and Methods

We conducted our experiments on inpatient mortality prediction. We followed these steps: 1) collecting discharge notes from medical records in the dataset; 2) applying the clinical NLP pipelines to extract concepts from clinical notes; 3) extracting key phrases using the TF-IDF (Term Frequency - Inverse Document Frequency) method, selecting concepts with positive scores, and eliminating others; 4) creating sequences of selected key phrases; 5) generating embedded sequences through different document representation models; 6) evaluating the performances of various embedding models. Fig. 1 demonstrates the whole workflow of the study.

### 3.1    Processing model

The proposed workflow includes the steps shown in Fig. 1. The primary database contains patients' clinical texts. From a variety of textual information, discharge sheets were selected for processing since they tend to include a summary of the most essential information concerning the patients' medical history and current hospitalization. This is especially useful for patients undergoing multiple treatments, as the outcome of each treatment is reported in this summary.

Each of the discharges notes, consists of several components or sections (25). However, there is significant variation in sections and the descriptive phrases in section headers. This challenge in the NLP of clinical notes is named flexible formatting (37, 40). After sectionizing each note by medspaCy (26), we follow the next step, which is named-entity recognition (NER). In this step, the clinical text processing was performed through data cleansing, NER, and entity linking stages. Through NER step, biomedical entities were identified and aligned to knowledge bases, promoting decision-making efficiency (2) and reducing clinical terminology diversity. A key goal of entity disambiguation and entity linking is to give each entity in the text a unique identifier. By using entity linkage, entities extracted from the text can be mapped to corresponding unique identifiers from a knowledge base, such as the Unified Medical Language System (UMLS).

The input of the model is an unstructured text, and its output is a list of all extracted concepts and the properties related to those concepts. This list is referred to as a bag of concepts hereinafter.

There are repetitive concepts in a bag of concepts. Following the removal of repetitive concepts, the remaining unique concepts of each patient were used as inputs in the different representation models. At this step, different representation models were used to generate an embedding vector of the text resulting from the connection of concepts related to each patient. For each representation model, a database containing the patient identifier and the embedding vector generated for them by that representation model was created.

These datasets were evaluated in an in-hospital death prediction problem such that the input of the model predicted the patient representation vector, and the goal was to predict the patient's death status. The XGBoost algorithm was used to predict the status of death.

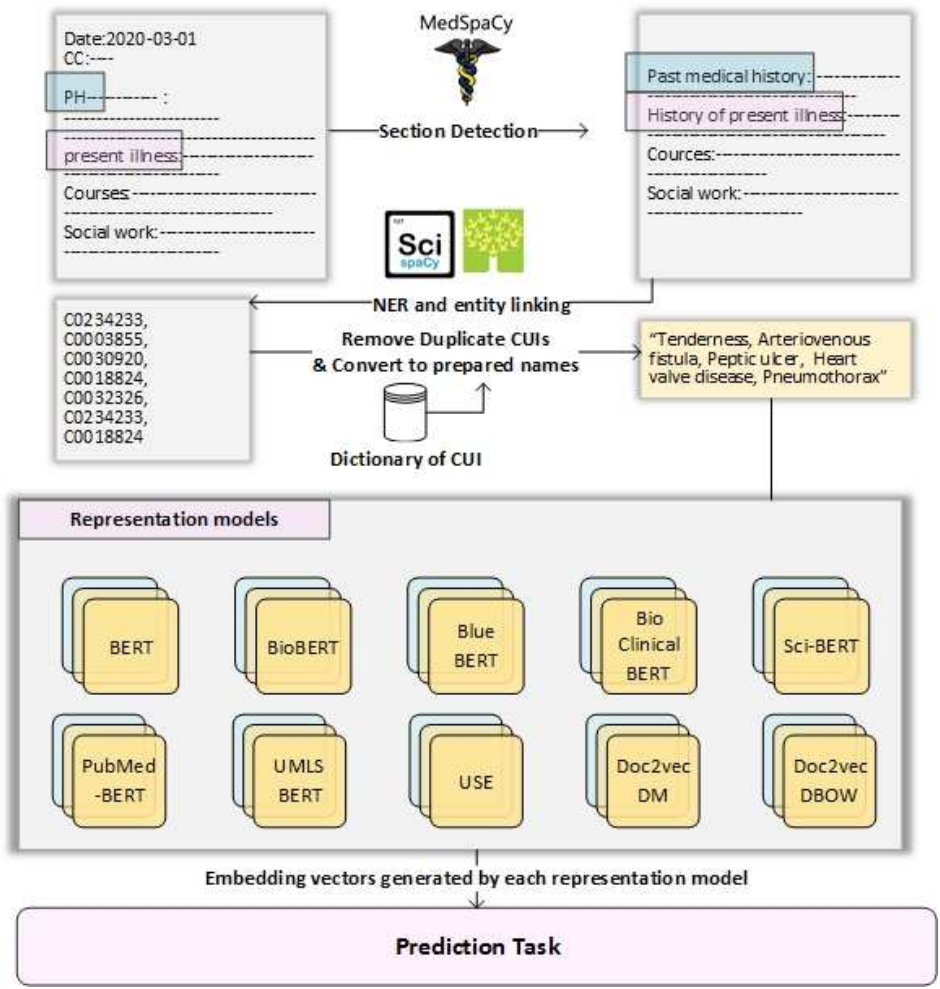

Fig. 1 Workflow of the methodology

### 3.2 Dataset

We used a subset of the MIMIC-III, a large, freely-accessible database comprising de-identified health-related data from more than forty thousand patients admitted to one medical center between 2001 and 2012. This subset consists of the extracted discharge summary of the first visits of 1610 patients (27)[5]. This set of EMRs was randomly selected from the MIMIC dataset (28).

While processing the summary files related to these patients, first, the different sections in the form were identified, and then the concepts extracted from the sections below were defined as following (29,30):

- **Past medical history**: A patient's health status prior to presenting their problem
- **History of present illness**: A description of the development of the patient's present illness
- **Chief complaint**: A concise statement describing the reason for the encounter
- **Family history**: Past occurrences (of a medical or mental health condition) in family members or past incidences (of a type of behavior) by family members
- **Physical exam**: An examination of the bodily functions and condition of an individual

The patient death status was selected from the admission table of the MIMIC dataset.

### 3.3 Clinical Text Preprocessing

A large amount of the information in an EMR is unstructured free text (37). Clinical text is written by various professionals such as physicians, nurses, physiotherapists, and psychologists. It is often written under time pressure and contains misspellings, non-standard abbreviations and jargon, and incomplete sentences. The use of general natural language processing (NLP) methods is insufficient for clinical texts (31). Accordingly, the following steps were followed for the processing:

There are sections in a patient's discharge sheet that contain valuable information about their current and past medical history. In this research, a collection of discharge sheets that have not been processed is called FullNotes. In the first step, it was necessary to identify the more valuable sections of the discharge sheets and separate them from other sections. The challenge

---

at this stage was that the headers or titles of these sections were not the same throughout the dataset, and different headers were used to refer to a specific section. This challenge in NLP of clinical notes is categorized as flexible formatting and named variable formatting (1,32). This problem was handled using the MedSpacy library (24).

A study on the data quality of the discharged sheets showed that 36 different information sections could be recorded. However, a limited number of these sections were completed in all EMRs (33). After identifying the sections within a discharge sheet and mapping the section headers to standard headers, a number of more important sections were selected in consultation with a physician, and their text was passed to the next step. In the next step, the type of each phrase was identified in terms of being negative/positive and disease/chemical. In the last step, the identified phrases were mapped with preferred names of standard concepts in the UMLS database. Therefore, for each identified phrase, we had a standard name (preferred name) and a unique code (concept unique identifier (CUI)). This step is known as entity linking. The libraries used in the implementation of different stages of the process are listed in Table 1.

Some of these concepts were repeated more than once. Duplicate CUIs were removed at this stage. It should be noted that the processing was first done on the CUIs, and in the last step, the remaining CUIs were converted to their preferred names in the knowledge base. This was done using a dictionary that had been produced from all positive and negative concepts in the corpus. This was to preserve the phrase attributed to the concepts whose standard names consisted of several words, some of which were common between different concepts. In addition, negative expressions would appear in the final output with a "NOT" in front of their name.

A summary text for each patient was created by connecting the preferred names of the remaining concepts. The final dataset contained each patient's ID and summary text, which are referred to as a unique concept (UC) herein. The generated summary text was used as the input of the next step to generate the embedded vector by the representation models.

**Table 1. the libraries used in the implementation of different stages of the process**

| Processing step | Tools/Python libraries | Setting | Ref |
|---|---|---|---|
| Section detection | MedspaCy | nlp.pipe_name : 'medspaCy_sectionizer' | (24) |
| Negation detection | Negspacy | | (34) |
| Concept mention detection, concept encoding | scispaCy | a Python library with en_ner_bc5cdr_md model | (17) |
| XGBClassifier | XGBoost | | (35) |
| Doc2Vec embedding model | Gensim | vector_size = 200, window_size= 5, sampling_threshold = 1e-5, negative_size = 5, alpha = 0.025, minimum_alpha = 0.0001, min_count =5, training_epochs = 10, DM=0 and1,hs=1(for hierarchical softmax) and negative =1 (for negative sampling) | (36) |
| Transformer-based embedding model | Hugging Face | | (37) |
| NearMiss | Imblearn.under_sampling | | (38) |

## 3.4 Representation

Different representation models were used to generate the embedding vector. These models can be classified into two groups depending on being transformer-based or not. Table 2 provides the information about the models.

**Table 2. The used document representation models**

| Model type | Model | Vocabulary | Domain-specific data | Year | Ref |
|---|---|---|---|---|---|
| Transformer-based | BioBERT | BERT-BASE | PubMed abstracts (PubMed) PubMed Central (PMC) | 2020 | (14) |
| | BLUEBERT | BERT-BASE | PubMed abstracts, clinical notes from the MIMIC III dataset | 2019 | (39) |
| | Bio+Clinical | BioBERT BERT-BASE | All MIMIC III, Only the discharge summaries in MIMIC III | 2020 | (40) |
| | SCI-BERT | SciVocab | Papers from the biomedical domain and computer science | 2019 | (41) |
| | PubMed-BERT | BERT-BASE | PubMed (abstracts and full biomedical articles) (3.1B words) | 2021 | (42) |
| | UMLS-BERT | Bio+Clinical-BERT | Patient notes and diagnostic test reports from the MIMIC III | 2020 | (43) |
| | BERT | BERT-BASE | - | 2018 | (44) |
| Other | USE | - | - | 2018 | (45) |
| | Doc2Vec | - | - | 2014 | (46) |

## 3.5  Downstream task

The generated representation vectors were used as the input of a classification model for predicting in-hospital deaths. The issue of predicting mortality from data recorded in EMRs is one of the common downstream tasks in many previous studies (47,48). This task can be defined as in-hospital or post-hospital mortality prediction. Identifying patients who are more likely to die allows for providing medical services on time to reduce their risk of death. According to the available dataset, this study implemented the in-hospital mortality prediction as a binary classification problem.

The XGBoost algorithm (35) was used to design the classification model. We produced separate datasets based on the representation models. Each dataset contained the patient identifier and the embedding vector generated by the model. The embedding vector was the input to the XGBoost algorithm. The target field was the patient's death status at the same admission. XGBoost is an ensemble method to combine decisions from multiple models for improved overall accuracy. The XGBoost algorithm builds base estimators sequentially and tries to reduce the bias of the combined estimator through gradient boosting.

## 4    Evaluation

The experiments were designed based on the following research questions:

1. What is the effect of the type of input generated from unstructured data on the performance of a representation model?

2. What is the difference between the performance of different representation models with the same input?

3. Changing which of the two factors, i.e., the input type or the representation algorithm, has a more significant impact on the final performance of the model?

We used three types of inputs to make the evaluation. These inputs were given to ten different representation models, and their generated embedding vectors were used in the downstream prediction task. The inputs were defined as follows:

1. **The Full Notes set**: This version comprises the entire text of case summaries after a standard text cleansing. The original texts are first broken down into 20 smaller parts for conversion into representation. Then the embedding vectors obtained for different parts are averaged.

2. **The Bag-of-Concept (BOC) set**: This version consists of all clinical concepts extracted from the text of case summaries. In this version, clinical concepts extracted from each clinical note are used to construct a sequence of concepts in the order in which they appear in the original text.

3. **Unique Concepts (UC)**: This version includes a set of unique concepts extracted from selected sections of the discharge sheets.

Table 1 shows the statistical information of these inputs. The three versions of the inputs were fed to the document representation models listed in Table 2.

**Table 3 Statistical information of three versions of the inputs**

|  | Full Notes len | Bag-of-Concept (BOC) len | Unique Concepts (UC) len |
|---|---|---|---|
| **Count** | 1528 | 1528 | 1528 |
| **Mean** | 149964 | 532 | 316 |
| **Std** | 28551 | 314 | 167 |
| **Min** | 100020 | 0 | 0 |
| **25%** | 125328 | 315 | 199 |
| **50%** | 151049 | 475 | 294 |
| **75%** | 173826 | 686 | 408 |
| **Max** | 199999 | 2273 | 1195 |

The evaluation was performed through K-fold cross-validation with K=2, and each model run 20 times. The same two folds were used throughout all the experiments. We took each fold as the test set and the remaining fold as the training set. Each ML algorithm learned on the training set and outputted its predictions for the test set, where evaluation metrics were calculated.

In this study, the XGBoost model was used for classification. Since the input data were imbalanced in terms of fatality, the near-miss method was used to under-sample the data (49). In transformer-based models, the last hidden state of the [CLS] token was used as the representation of the entire input sentence. The input vector size was set to 768 for transformer-based models and PV-DM and PV-DBOW for Doc2Vec and USE models, respectively. The effectiveness of the proposed representation method was evaluated in a downstream prediction task. The evaluation metrics are described below.

## 4.1 Evaluation metrics

**Precision (P)**: is defined as the number of true positives ($T_p$) over the number of true positives plus the number of false positives ($F_p$).

$$P = \frac{T_p}{T_p + F_p} \tag{1}$$

**Recall (R):** is defined as the number of true positives ($T_p$) divided by the sum of the number of true positives and number of false negatives ($F_n$).

$$R = \frac{T_p}{T_p + F_n} \tag{2}$$

**(F1) score**: is defined as the harmonic mean of precision and recall.

$$F_1 = 2 \times \frac{P \times R}{P + R} \tag{3}$$

**Accuracy**: is defined as

$$Accuracy = \frac{Sensitivity + Specificity}{2} \tag{4}$$

## 4.2 In-hospital death prediction

This section presents the results of in-hospital death prediction using the embedding models listed in Table 1 on the inputs derived from patient case summaries.

The following can be concluded from the results presented in Table 4 and Table 5.

**TABLE 4 - Results of the prediction model (Recall and Precision )**

| Representation models | Recall | | | Precision | | |
|---|---|---|---|---|---|---|
| | BOC | Full Notes | UC | BOC | Full Notes | UC |
| **BERT-BASE** | 0.48 | 0.72 | 0.76 | 0.53 | 0.81 | 0.83 |
| **BIO-BERT** | 0.52 | 0.75 | **0.84** | **0.94** | 0.82 | 0.86 |
| **BIO-CLINICAL-BERT** | 0.54 | 0.74 | 0.82 | 0.61 | 0.79 | 0.84 |
| **BLUEBERT** | 0.71 | 0.73 | 0.75 | 0.82 | 0.81 | 0.84 |
| **DOC2VEC-DM-0** | 0.73 | 0.73 | 0.71 | 0.77 | 0.71 | 0.78 |
| **DOC2VEC-DM-1** | **0.79** | **0.80** | 0.71 | 0.82 | **0.85** | 0.76 |
| **PUBMED-BERT** | 0.44 | 0.71 | 0.83 | 0.60 | 0.77 | 0.86 |
| **SCI-BERT** | 0.71 | 0.77 | 0.76 | 0.81 | 0.83 | **0.87** |
| **UMLSBERT** | 0.71 | 0.72 | 0.81 | 0.80 | 0.81 | 0.86 |
| **USE** | 0.69 | 0.71 | 0.78 | 0.93 | 0.79 | 0.81 |

**Table 5 - Results of the prediction model (Accuracy and F1)**

| Representation models | Accuracy | | | F1 | | |
|---|---|---|---|---|---|---|
| | BOC | Full Notes | UC | BOC | Full Notes | UC |
| **BERT-BASE** | 52.15 | 77.35 | 79.92 | 0.51 | 0.77 | 0.80 |
| **BIO-BERT** | 73.77 | 78.69 | **85.04** | 0.72 | 0.79 | **0.85** |
| **BIO-CLINICAL-BERT** | 59.27 | 77.08 | 83.38 | 0.59 | 0.77 | 0.83 |
| **BLUEBERT** | 77.35 | 77.42 | 80.31 | 0.77 | 0.77 | 0.80 |
| **DOC2VEC-DM-0** | 75.73 | 71.52 | 75.50 | 0.76 | 0.71 | 0.75 |
| **DOC2VEC-DM-1** | 80.81 | **82.73** | 74.23 | **0.81** | **0.83** | 0.74 |
| **PUBMED-BERT** | 55.42 | 74.77 | 84.58 | 0.54 | 0.75 | 0.85 |
| **SCI-BERT** | 77.04 | 80.08 | 82.27 | 0.77 | 0.80 | 0.82 |
| **UMLSBERT** | 76.12 | 77.46 | 83.31 | 0.76 | 0.77 | 0.83 |
| **USE** | **81.27** | 75.50 | 79.77 | **0.81** | 0.75 | 0.80 |

In Figures 2 and 3, the results are visually represented. Based on these results, a number of important conclusions can be drawn:

1. All transformer-based models achieved their best performance level when run with UC inputs.

2. The model offering the best results with the Full Notes inputs was Doc2Vec-PM (Doc2Vec-DM-DBOW).

3. The model achieving the best results with UC inputs in terms of (F1) score and accuracy was the transformer-based Bio-BERT ((F1) score = 0.850; accuracy = 0.852). The models with the next highest performance level with UC inputs were PubMed-BERT, followed by UMLS-BERT and Bio+Clinical-BERT.

4. According to Figure 2, all transformer-based models with BOC input had a lower average F1 metric than other input types.

5. In light of the superior results obtained from UC input, it proves the effectiveness of removing duplicate concepts from embedding vectors.

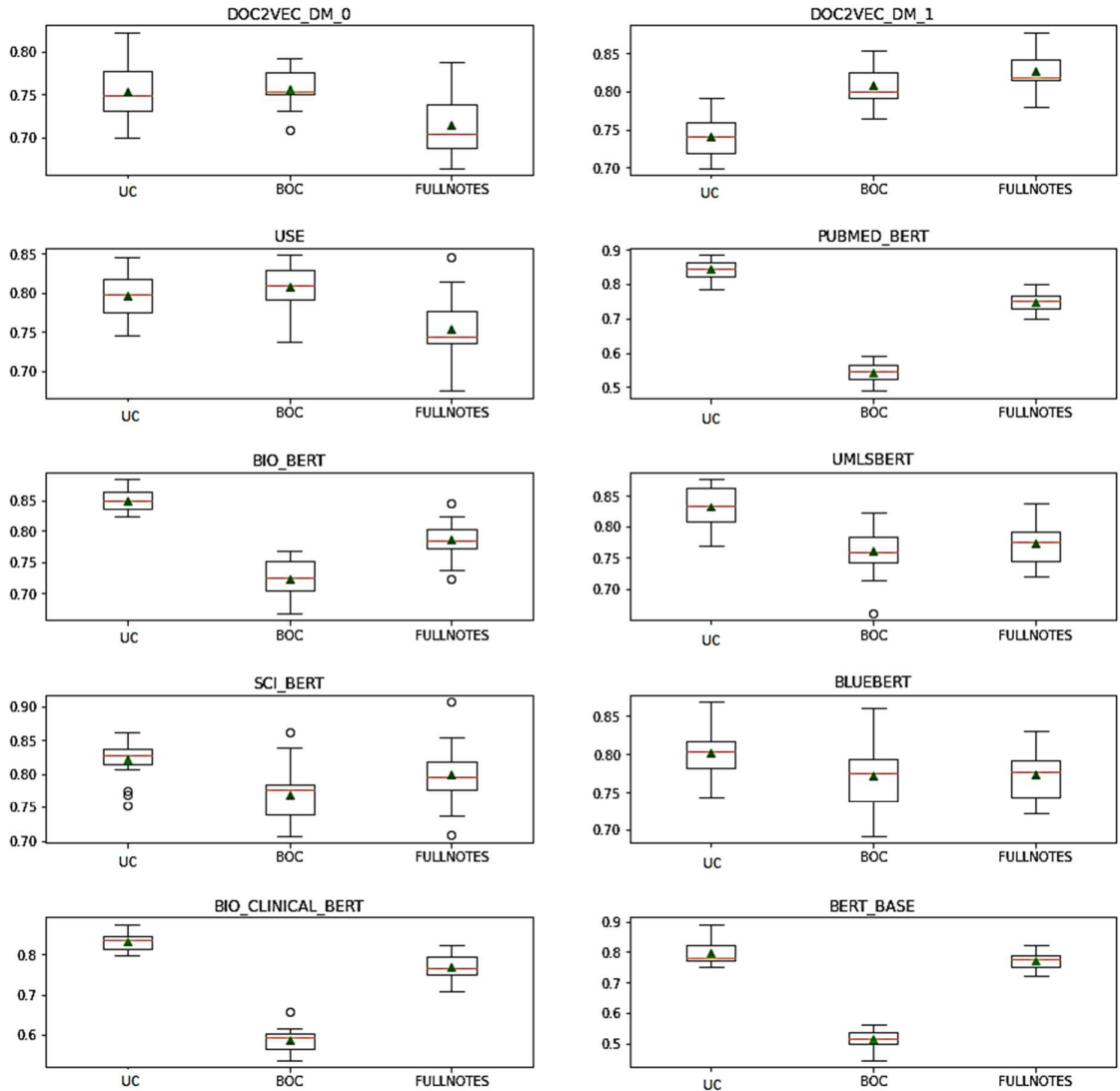

Fig. ٣ The difference between the performances of representation models in F1 score with three types of inputs

According to Figure 3, PubMed-BERT, BERT-BASE, and Bio+Clinical-BERT models were the most changeable when receiving different inputs. In contrast, models SCI-BERT , BLUEBERT  and UMLSBERT are less affected by changing the type of input.

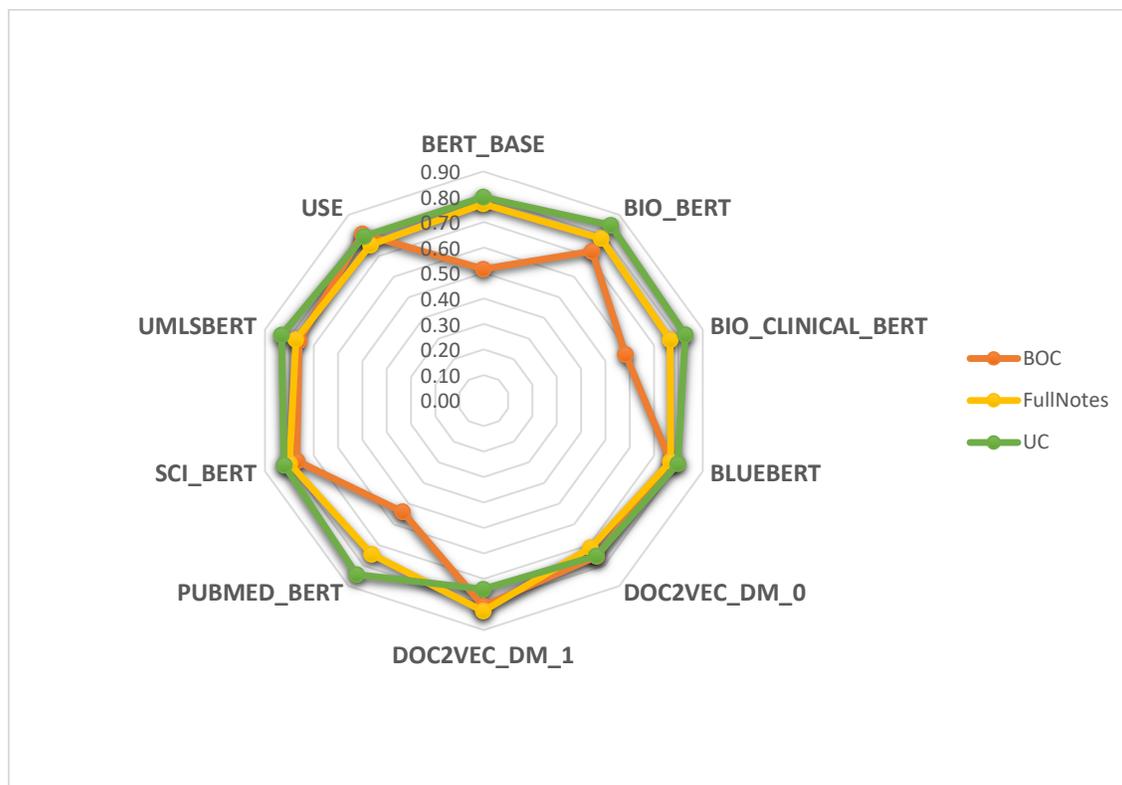

Fig. 3. The accuracy chart of the representation model with different inputs

As shown in Figure. 4, all transformer-based models performed better with UC inputs than with BOC inputs (the performance difference is marked with orange color). The values given in the figure indicate that accuracy improved when the models were fed with unique concepts (UC) instead of all extracted concepts (BOC). The most remarkable improvements were provided by PubMed-BERT, which achieved 30% performance improvement by using unique concepts rather than all extracted concepts.

It can also be seen that all models except Doc2Vec-Dm-DBOW had better performance with unique concept inputs than with FullNotes inputs (the performance difference is marked with blue color). In other words, using unique concepts as the model input rather than full texts improved the performance of most models in terms of accuracy.

Fig. 4 shows that using FullNotes inputs instead of BOC inputs, which are shown with blue bars in the plot, led to better results with all transformer-based models (green bars), with the most significant performance change belonging to BERT-BASE (26%) and PubMed-BERT (21%).

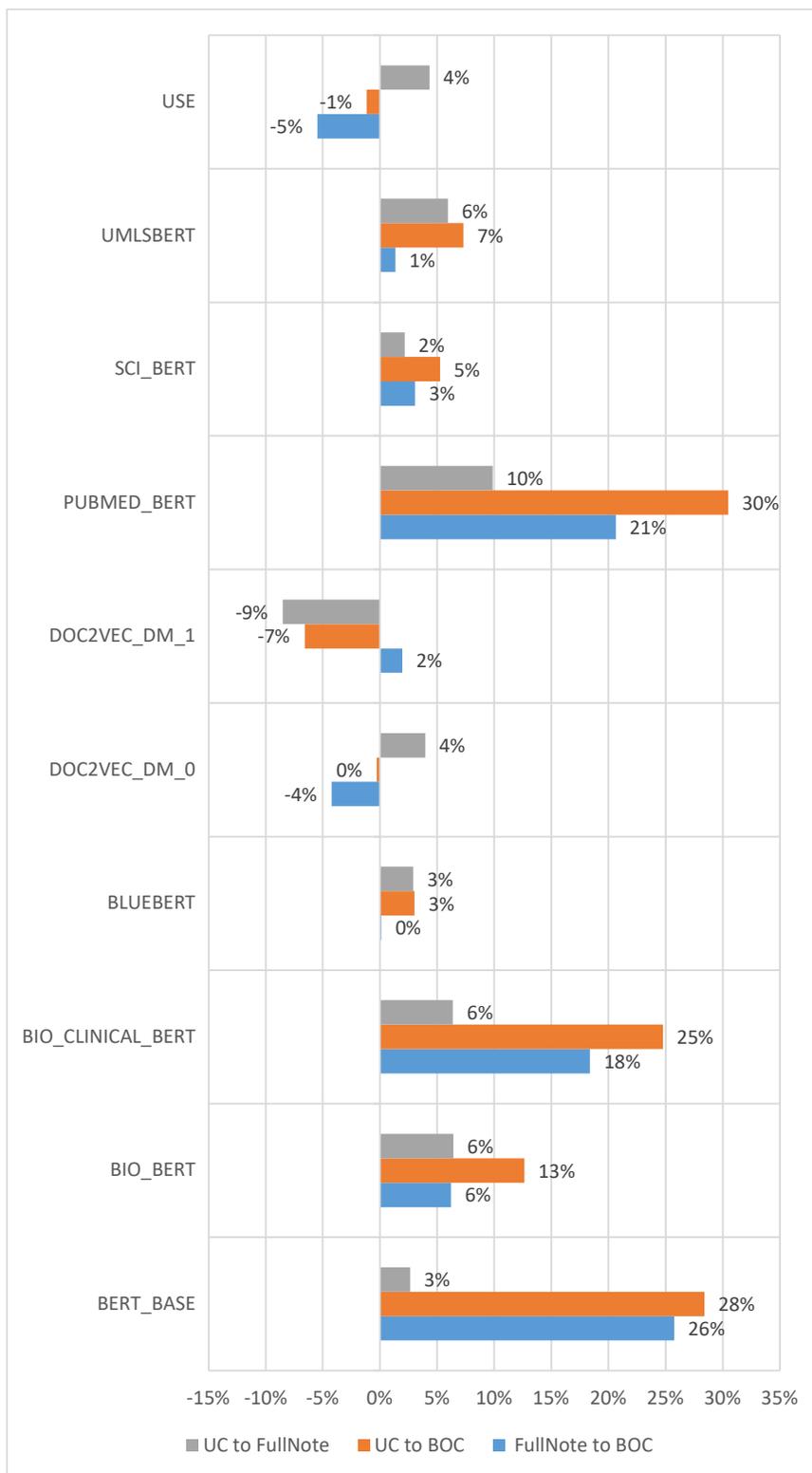

Figure 5 A comparison between the representation models performance on different inputs types in terms of accuracy.

## 5    Conclusion

Given the widespread use of EMRs, massive amounts of clinical data are generated every day. Ongoing improvements in artificial intelligence and access to processing resources have provided a cost-effective alternative way to classify and screen patients using EMR-generated data. This study investigated the effectiveness of clinical text representation methods using new transformer-based models that have been developed in recent years.

Since the data recorded in clinical notes could be influenced by a variety of factors such as physicians' knowledge and competence, extracting key clinical concepts from these texts and mapping them with standard terminologies can be a great step toward identifying similar cases from texts. However, a great number of concepts extracted from clinical texts tend to be essentially irrelevant and useless for the purpose of text analysis. In this study, instead of using the bag-of-concepts, first, key concepts were extracted using a classical statistical method, and then the extracted concepts were used to construct summary phrases to be used as inputs in the representation methods.

The results showed that using key phrases instead of bag-of-concepts improved the performance of all transformer-based models tested in the study. This improvement can be attributed to the elimination of irrelevant and repetitive concepts. Another finding of this study was the poorer performance of all models when they were run with bag-of-concepts compared to when they were given Full Notes. This shows that extracting all concepts, including irrelevant ones, cannot compensate for the loss of word relationships.

Another finding of this study was the differences observed in the performance levels of different BERT-BASE-based models when run with different input types. The best results were obtained with Bio-BERT and PubMed-BERT when using unique concepts as input. This better performance can be attributed to the use of scientific articles in the pre-training phase of these two models. The third best performance with unique concepts as inputs (with only a 1% difference from the second-best performance) belonged to UMLS-BERT, which has been pre-trained with a UMLS knowledge base.

This study also had several limitations. The first limitation was the mapping of clinical concepts. Although the various tools that can be used for this purpose return more than one concept for each clinical term, in this study, only the concept with the highest score was picked for use in later steps. In future works, we will focus on extracting key concepts and using the different methods that exist in this area.

## 6    Availability of data and materials

The dataset supporting the conclusions of this article is available in phenotyping

Repository, https://github.com/sebastianGehrmann/phenotyping.git.